\documentclass[10pt,twocolumn,letterpaper]{article}

\usepackage{iccv}              

\usepackage{graphicx}
\usepackage{amsmath}
\usepackage{amssymb}
\usepackage{booktabs}

\usepackage{times}
\usepackage{epsfig}
\usepackage{graphicx}
\usepackage{amsthm}
\usepackage{caption}
\usepackage{multirow}
\usepackage{makecell}
\usepackage{comment}
\usepackage[table,xcdraw,dvipsnames]{xcolor}
\usepackage{algorithm}
\usepackage{algpseudocode}
\usepackage{pifont}
\usepackage{enumitem}
\usepackage{bbm}

\definecolor{iccvblue}{rgb}{0.21,0.49,0.74}
\usepackage[pagebackref,breaklinks,colorlinks,allcolors=iccvblue]{hyperref}

\def\paperID{5341} 
\def\confName{ICCV}
\def\confYear{2025}

\title{DAMap: Distance-aware MapNet for High Quality HD Map Construction}

\author{Jinpeng Dong, Chen Li, Yutong Lin, Jingwen Fu, Sanping Zhou, Nanning Zheng\thanks{Corresponding author.}\\
State Key Laboratory of Human-Machine Hybrid Augmented Intelligence,\\ National Engineering Research Center for Visual Information and Applications,\\ Institute of Artificial Intelligence and Robotics, 
Xi'an Jiaotong University\\
{\tt\small \{djp1235a, edward82, yutonglin, fu1371252069\}@stu.xjtu.edu.cn} \\ 
{\tt\small \{spzhou, nnzheng\}@xjtu.edu.cn}
}

\begin{document}
\maketitle
\begin{abstract}
Predicting High-definition (HD) map elements with high quality (high classification and localization scores) is crucial to the safety of autonomous driving vehicles. However, current methods perform poorly in high quality predictions due to inherent task misalignment. Two main factors are responsible for misalignment: 1) inappropriate task labels due to one-to-many matching queries sharing the same labels, and 2) sub-optimal task features due to task-shared sampling mechanism. In this paper, we reveal two inherent defects in current methods and develop a novel HD map construction method named DAMap to address these problems. Specifically, DAMap consists of three components: Distance-aware Focal Loss (DAFL), Hybrid Loss Scheme (HLS), and Task Modulated Deformable Attention (TMDA). The DAFL is introduced to assign appropriate classification labels for one-to-many matching samples. The TMDA is proposed to obtain discriminative task-specific features. Furthermore, the HLS is proposed to better utilize the advantages of the DAFL. We perform extensive experiments and consistently achieve performance improvement on the NuScenes and Argoverse2 benchmarks under different metrics, baselines, splits, backbones, and schedules. Code will be available at https://github.com/jpdong-xjtu/DAMap.

\end{abstract}    
\section{Introduction}
\label{sec:intro}

High-definition (HD) map is an important component developed specifically for autonomous driving vehicles.
Traditional offline HD maps are constructed using SLAM-based methods \cite{zhang2014loam,shan2018lego} that require manual annotation on the Lidar point clouds. The offline pipeline is complex, high-cost, and difficult to keep up-to-date frequently.
To address these challenges, there has been a growing interest in the development of online HD map.
\begin{figure}[t]
\setlength{\abovecaptionskip}{0.1cm}
\setlength{\belowcaptionskip}{-0.2cm}
 \centering
 \subcaptionbox*{}{\includegraphics[width=0.99\linewidth]{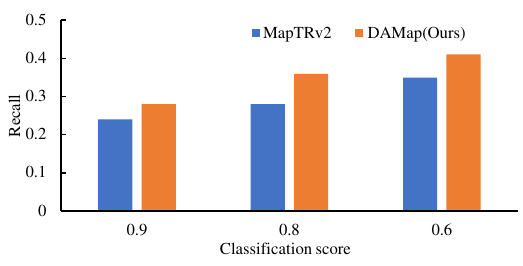}} \\
\subcaptionbox*{}{\includegraphics[width=0.98\linewidth]{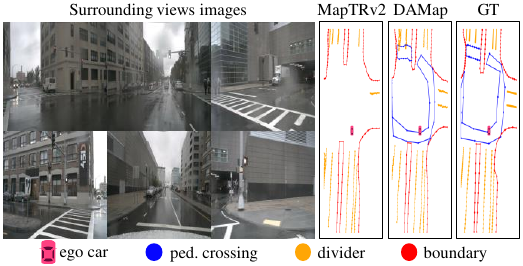}}
 \caption{Top: The analysis about recall at 0.5m threshold (high localization). 
Even with the 0.6 classification threshold, the recall is only $35\%$. This indicates that the baseline performs poorly in high quality predictions. Our method improves the recall of high quality predictions at each score threshold.
Bottom: A visualization example with a 0.4 classification threshold. At this threshold, the baseline model fails to predict pedestrian crossings, which is dangerous for autonomous driving vehicles.
}
 \label{fig:issue}
\end{figure}
Inspired by DETR \cite{carion2020end,zhudeformable,zhangdino,lin2023plaindetr,dong2024augdetr} and BEV perception \cite{li2022bevformer,huang2021bevdet,philion2020lift}, 
MapTR \cite{liaomaptr} designs a transformer-based pipeline with permutation-equivalent modeling and hierarchical query embedding for more efficient HD map construction. MapTRv2 \cite{liao2024maptrv2}builds upon the MapTR framework by incorporating decoupled self-attention to reduce memory consumption and the one-to-many matching to speed up convergence.

Predicting map elements with high classification and localization quality is crucial to the safety of autonomous driving vehicles. 
Although MapTRv2 \cite{liao2024maptrv2} yields promising performance gains, the performance of high quality predictions is still poor, as analyzed in Figure \ref{fig:issue}.
We argue that two inherent defects in current methods limit high quality predictions as follows:

\textbf{Inappropriate task labels} is that one-to-many matching samples share the same labels easily leads to the misalignment results of high classification and low localization.
When one-to-many matching branches are introduced, each GT instance corresponds to several different candidate samples. 
For example, each GT map element instance in the MapTRv2 \cite{liao2024maptrv2} corresponds to 7 candidate samples. When using loss backpropagation training, different candidate samples are given the same classification labels, ignoring the differences in localization of different samples. This way allows low localization samples to output high classification scores.
Many results with high classification but low localization will result in poor performance.
Based on the above observations, we propose to assess the localization quality of these candidate samples and assign proper labels based on this assessment. In this way, the potential of one-to-many branching can be better exploited, and low quality predictions can be suppressed. 

\textbf{Sub-optimal task features} are that task-shared features may cause inherent conflicts between tasks, leading to sub-optimal task outputs.
Cross-attention layer (deformable attention) is used to extract the instance features through the interaction between queries and image features. In the current map construction framework, the classification and localization branches use the same instance features. However, classification and localization tasks usually have different feature preferences. Conventional deformable attention does not extract task-specific features. This is not conducive to obtaining more accurate results for both tasks. 
Typically, utilizing task modulated methods tends to obtain more flexible feature representations. 
So, we argue that deformable attention in the decoder can be modulated to extract discriminative task-specific features. 

In this paper, we propose the DAMap, a simple yet effective HD map construction method that uses three components to obtain high quality predictions. Specifically, Distance-aware Focal Loss (DAFL) is proposed to assign appropriate classification labels for one-to-many matching samples by introducing the localization quality of candidate samples into the classification loss. The DAFL utilizes likelihood estimation to convert localization loss into localization quality. Task Modulated Deformable Attention (TMDA) is introduced to obtain discriminative task-specific sampling features by learning task-specific sampling weights in the cross-attention layer. Furthermore, to better utilize the advantages of our DAFL, we propose a Hybrid Loss Scheme (HLS) to use Focal Loss \cite{lin2017focal} and our DAFL in different decoders during the training stage.
We perform extensive experiments to validate the effectiveness of DAMap on the NuScenes \cite{caesar2020nuscenes} and Argoverse2 \cite{wilson2argoverse} benchmark. Our method consistently achieves performance improvement. 
We summarize our contributions as follows:
\begin{itemize}
\item We analyze two inherent defects in the current methods, inappropriate classification labels and sub-optimal task features, which limit the predictions with high classification and localization quality.
\item A new online vectorized HD map construction method named DAMap is proposed to address these defects with Distance-aware Focal Loss, Task Modulated Deformable Attention, and Hybrid Loss Scheme.
\item We validate the proposed DAMap on the NuScenes and Argoverse2 benchmarks under different metrics, baselines, splits, backbones, and schedules, and it consistently yields performance improvement.
\end{itemize}

\section{Related Work}
\begin{figure*}[ht]
\setlength{\abovecaptionskip}{0.1cm}
\setlength{\belowcaptionskip}{-0.2cm}
 \centering
\includegraphics[width=0.98\linewidth]{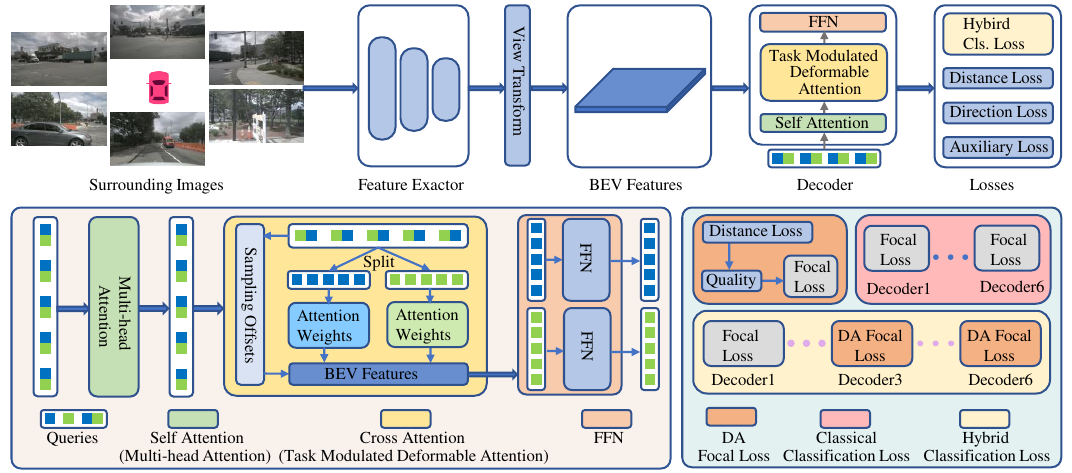}
 \caption{The pipeline of the proposed DAMap. Taking surrounding images as inputs, a shared backbone as the feature extractor first extracts image features; The extracted image features are then transformed to BEV space as BEV features; The transform decoder finally achieves the interaction between queries and BEV features to predict instances of map elements. These predictions are used to calculate the loss for model optimization in the training phase. }
 \label{fig:pipeline}
\end{figure*}
\subsection{HD Map Construction}
HDMapNet \cite{li2022hdmapnet} proposes to group pixel-level segmentation results into vectorized representations. 
VectorMapNet \cite{liu2023vectormapnet} further proposes a two-stage coarse-to-fine framework with the auto-regressive decoder to sequentially generate points. 
Inspired by DETR\cite{carion2020end}, MapTR \cite{liaomaptr} designs an end-to-end DETR-like pipeline with permutation-equivalent point set modeling and hierarchical query embedding.
MapTRv2 \cite{liao2024maptrv2} builds upon the MapTR by incorporating decoupled self-attention to reduce memory consumption and the one-to-many matching branch to speed up the convergence.
Piece-wise Bézier curve modeling is introduced into BeMapNet \cite{qiao2023bemap}, and PivotNet \cite{ding2023pivotnet} utilizes dynamic pivot-based points to represent map elements.
MGMap \cite{liu2024mgmap} utilizes the guidance of learned masks to enhance element features in detail. 
HIMap \cite{zhou2024himap} proposes the hybrid representation learning with a point-element interaction module and a point-element consistency constraint scheme. MapQR \cite{liu2024mapqr} replaces the point-element query with the proposed scatter-and-gather query to address content conflicts that points of the same map element may have different content features. In addition, StreamMapNet \cite{yuan2024streammapnet} proposes a long-sequence temporal modeling pipeline for complex scenarios and broader perception ranges. 
In contrast to these works, our focus lies on task labels and features for high quality predictions. In addition, we conducted experiments using these SOTA methods as the baseline models, and our method is complementary to theirs.

\subsection{High Quality Learning}
Although high quality learning is less focused on HD map construction, there exists some works \cite{jiang2018iounet,li2020generalized,zhang2021varifocalnet,dong2025novel,cai2023align,pu2024rank} to improve high quality predictions for other tasks. For example, IoU-Net \cite{jiang2018iounet} introduces an auxiliary Intersection over Union (IoU) head to output IoU scores for post-processing. GFL \cite{li2020generalized} designs a quality focal loss to obtain the joint representation of classification scores and IoU scores. VFNet \cite{zhang2021varifocalnet} proposes a novel IoU-based classification loss and a refinement scheme to obtain more accurate results. Since these methods achieved success in the convolutional era, some recent works like Stable-DINO \cite{liu2023stable}, Align-DETR \cite{cai2023align}, and Rank-DETR \cite{pu2024rank} introduce these ideas into the DETR framework. In addition to assigning an appropriate label, there are some works \cite{wu2020doublehead,song2020tsd,dong2022construct,dong2026monoa,zhang2023decoupleddetr} that explore task feature enhancement. Double-Head R-CNN \cite{wu2020doublehead} uses two separate branches to decouple the sibling task head. TSD \cite{song2020tsd} proposes a spatial disentanglement module to generate different proposals for different tasks. For the DETR \cite{carion2020end} framework, DESTR \cite{he2022destr} and Decoupled DETR \cite{zhang2023decoupleddetr} utilize two cross-attention (multi-head attention) modules for different tasks. Although these methods have been successful, their efficacy in the HD map construction requires further validation. 
Furthermore, our main contribution, assigning an appropriate label without explicit IoU representation, designing the task modulated deformable attention, and hybrid using loss in different decoders, has not yet been explored in previous works.
\section{Method}

\subsection{Framework Overview}
Taking multi-view RGB images as inputs, our framework is to predict instances of map elements. 
The overall pipeline of our DAMap is shown in Figure \ref{fig:pipeline}. Giving multi-view RGB images as input, a shared backbone as features extractor is first used to extract image features. Then, these extracted multi-view image features are transformed to BEV space with a pre-defined size $H\times W$ through the view transform module. The generated BEV features are denoted as $\mathbf{F} \in \mathbb{R}^{C\times H \times W}$, where $C$ represents the number of channels. Finally, the DETR-like transform decoder is used to achieve the interaction between queries and BEV features and to generate instances of map elements.
\subsection{Preliminary}
\textbf{Focal Loss} (FL) \cite{lin2017focal} is a classical classification loss designed to address extreme imbalance between positive and negative samples during the object detector training stage. It is also used as the classification loss in the HD map construction task. It can be simply formulated as:

\begin{equation}
	\begin{split}
       &\text{FL}(p,y)=\begin{cases}
       -(1-p)^\gamma\log(p),& \text{ \quad $ y=1 $ } \\
       -p^\gamma\log(1-p),& \text{ \quad $ y=0  $ }
\end{cases}\\
   \end{split}
   \label{fl_loss}
\end{equation}
where $y\in\{1,0\}$ indicates the ground truth class and $p\in[\,0,1]\,$ denotes the predicted classification probability score for the foreground class. $\gamma$ is the hyper-parameter to adjust the effect of the modulation factor. As shown in Equation \ref{fl_loss}, focal loss consists of a modulation scaling factor part $(1-p)^\gamma$ and a conventional cross entropy part $-\log(p)$. Specifically, the modulation scaling factor $(1-p)^\gamma$ and $p^\gamma$ correspond to the foreground and background classes, respectively. 

 \textbf{Decoder} is used to extract the features of map instances through interaction between queries and image features. The decoder consists of a self-attention layer, a cross-attention layer, and an FFN layer. Let $\mathbf{Q}$ denote the predefined queries and $\mathbf{F}$ denote the generated BEV features. The pipeline of the decoder can be simply formulated as:
\begin{equation}
  \hat{\mathbf{Q}}=\{\mathrm{SA}(\mathbf{Q}), \mathrm{CA}(\mathbf{Q},\mathbf{F}), \mathrm{FFN}(\mathbf{Q})\},
\end{equation}
where $\mathrm{SA}(\cdot)$ is the self-attention layer that performs interactions between different queries. $\mathrm{CA}(\cdot)$ is the cross-attention layer that extracts the instance features from the image features. In MapTRv2 \cite{liao2024maptrv2}, Deformable Attention is used to construct a cross-attention layer.  $\mathrm{FFN}(\cdot)$ is the feed-forward network layer that enhances the instance features. The normalization layers are omitted in the pipeline. 

After obtaining the instance features extracted by the decoder, the instance features are fed into the classification and localization task heads, consisting of linear layers, to output the predictions. The process can be formulated as:
\begin{equation}
  \mathbf{O}_{\mathrm{cls}}=\mathrm{H}_{\mathrm{cls}}(\hat{Q}), \mathbf{O}_{\mathrm{loc}}=\mathrm{H}_{\mathrm{loc}}(\hat{Q}),
\end{equation}
where $H_{cls}$ indicates the classification head, $H_{loc}$ indicates the localization head. The final predictions are usually obtained by cascading multiple decoders and task heads.
\begin{figure}[t]
\setlength{\abovecaptionskip}{0.1cm}
\setlength{\belowcaptionskip}{-0.3cm}
 \centering
\includegraphics[width=0.98\linewidth]{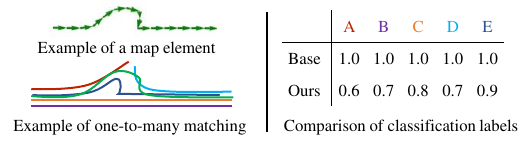}
 \caption{A simple example of inappropriate classification labels for one-to-many matching queries. The green line represents GT, and the other colored lines represent queries.}
 \label{fig:queries_labels}
\end{figure}

\subsection{Distance-aware Focal Loss}

Although the integration of Focal Loss can speed up convergence and yield promising performance gains, its modulation factor, considering only the classification score of samples, is not optimal. 
When one-to-many matching branches are introduced, each GT instance corresponds to several different candidate samples. However, these different candidate samples are given the same label of 1, which is not conducive to obtaining a more discriminative model, as shown in Figure \ref{fig:queries_labels}. We need to assess the localization quality of these candidate samples and assign appropriate labels based on this localization assessment. 

The localization loss can be used to indicate the localization quality, so we first introduce the localization loss (distance loss) as follows: 
\begin{equation}
   \mathcal{L}_{\rm{dist}} = \sum_{i=0}^{N-1}   \mathbbm{1}_{\{c_i\neq \varnothing\}}    \sum_{j=0}^{N_v-1} D_{\rm {L1}}(\hat{v}_{\hat{\pi}(i),j},  v_{i, \hat{\gamma}_i(j)}),
\end{equation}
where $i$ is the index of GT instance, $j$ is the index of point in GT instance, $\hat{\gamma}_i$ denotes the point-level matching result, $\hat{\pi}$ denotes instance-level matching result. $\hat{v}_{\hat{\pi}(i),j}$ represents each predicted point, $v_{i, \hat{\gamma}_i(j)}$ represents the assigned a GT point. The localization loss is obtained by computing the L1 distance between predicted points and assigned GT points.

The large range of values for the distance loss ($\mathcal{L}_{\rm{dist}} \in [\,0,+\infty]\,$) causes it not to be used as directly as the classification score ($p\in[\,0,1]\,$). To align the classification scores, we need to represent the distance loss as a probability to indicate the localization quality. The distance loss can be converted into a likelihood probability using maximum likelihood estimation theory, as follows:
\begin{equation}
   P_{dist}^i=e^{-\lambda\mathcal{L}_{\rm{dist}}^{i}},
\end{equation}
where $P_{dist}^i$ denotes the instance localization confidence to indicate the localization quality. When $\mathcal{L}_{\rm{dist}}^{i}=0$, $P_{dist}$ is equal to 1. When $\mathcal{L}_{\rm{dist}}^{i}$ tends to infinity, $P_{dist}$ tends to 0. $\lambda$ is a hyper-parameter to adjust the effect of loss.

After obtaining the localization probability, we can use the localization probability as continuous labels to assign more reasonable labels to different samples. Specifically, we replace the binary labels in Focal Loss with continuous labels and thus obtain the proposed Distance-aware Focal Loss (DAFL). The DAFL is defined as:
\begin{equation}
       \text{DAFL}(p,y)=-(y-p)^\gamma(y\log(p)+(1-y)\log(1-p)),
   \label{dafl_loss}
\end{equation}
where $y\in[\,0,1]\,$ denotes the localization confidence. As shown in Equation \ref{dafl_loss}, when the sample is a negative sample ($y=0$), the DAFL is the same as the raw Focal Loss. When the samples are positive, the localization confidence will be used as the classification label as a supervised signal so that the classification score can indicate the localization quality. 

Our proposed DAFL can mitigate the issue of the classification task not perceiving the localization quality, thus avoiding the output of predictions with high classification and low localization quality.

\subsection{Hybrid Loss Scheme}
MapTRv2\cite{liao2024maptrv2} typically stacks 6 decoder layers to output predictions using a cascade style. All decoder layers are supervised by classification and localization loss. The total classification loss can be simply formulated as:
\begin{equation}
   \mathcal{L}_{\rm{cls}} = \sum_{l=1}^{L}\text{FL}(p,y),
\end{equation}
where $L$ is the number of decoder layers.

To introduce our proposed DAFL into the classification loss, a straightforward way is to replace the Focal Loss in all decoders with our proposed DAFL. The total classification loss is converted as:
\begin{equation}
   \mathcal{L}_{\rm{cls}} = \sum_{l=1}^{L}\text{DAFL}(p,y).
\end{equation}

The random initialization of the query in MapTRv2 \cite{liao2024maptrv2} leads to poor localization results at the early stage of training, which may affect our proposed DAFL. 

To take full advantage of our proposed DAFL, we propose the hybrid loss scheme to combine the Focal Loss \cite{lin2017focal} with the proposed DAFL. Specifically, we use Focal Loss as the classification loss in the earlier decoder layer and DAFL as the classification loss in the later decoder layer. Focal Loss is independent of the localization quality of samples. Predictions of the later decoders tend to have a higher localization quality than the earlier ones due to the cascade property of the decoders. The total classification loss can be simply formulated as:
\begin{equation}
   \mathcal{L}_{\rm{cls}} = \sum_{l=1}^{L1}\text{FL}(p,y)+\sum_{l=1}^{L2}\text{DAFL}(p,y),
\end{equation}
where $L1$ is the number of decoder layers with Focal Loss, $L2$ is the number of decoder layers with DAFL. The sum of $L1$ and $L2$ is equal to $L$.
\begin{table*}[ht]
\setlength{\tabcolsep}{3.66pt}
\setlength{\abovecaptionskip}{0.1cm}
\setlength{\belowcaptionskip}{0.cm}
    \centering
    \scalebox{0.98}{
    \begin{tabular}{c| c| c|c c c |c|c c c |c} 
    \Xhline{1.5pt}
    \multirow{2}{*}{Methods} &
    \multirow{2}{*}{Reference} &
    \multirow{2}{*}{Epoch} &
    $\text{AP}_{ped.}$ & $\text{AP}_{div.}$ & $\text{AP}_{bou.}$ & mAP &
    $\text{AP}_{ped.}$ & $\text{AP}_{div.}$ & $\text{AP}_{bou.}$ & mAP \\
    \cline{4-11}
    & & & \multicolumn{4}{c|}{ $hard:\{0.2, 0.5, 1.0\}m$} & \multicolumn{4}{c}{ $easy:\{0.5, 1.0, 1.5\}m$} \\
    \hline
    HDMapNet \cite{li2022hdmapnet} &  ICRA22 &  30&  7.1 &  28.3 &  32.6  &  22.7 & 24.1 & 23.6 & 43.5 & 31.4\\ 
    PivotNet \cite{ding2023pivotnet} &  CVPR23 &  30 & 34.8 & 42.9 & 39.3 & 39.0 & 53.8 & 58.8 & 59.6 &  57.4 \\ 
    BeMapNet \cite{qiao2023bemap} &  CVPR23 &  30 & 39.0 & 46.9 & 37.8 & 41.3 & 57.7 & 62.3 & 59.4 &  59.8 \\
    MapTR \cite{liaomaptr} &  ICLR23 &  24 & 23.2 & 30.7 & 28.2 & 27.3 & 46.3 & 51.5 & 53.1 &  50.3 \\
    StreamMapNet \cite{yuan2024streammapnet} &  WACV24 &  30 & - & - & - & - & 61.7 & 66.3 & 62.1 &  63.4 \\
    MapTRv2 \cite{liao2024maptrv2} &  IJCV24 &  24 & 35.4 & 40.0 & 36.3 & 37.2 & 59.8 & 62.4 & 62.4 &  61.5 \\
    MapTRv2+MGMap \cite{liu2024mgmap} &  CVPR24 &  24 & - & - & - & - & 61.8 & 65.0 & 67.5 &  64.8 \\
    HIMap \cite{zhou2024himap} &  CVPR24 &  30 & 37.2 & 49.2 & 42.8 & 43.1 & 62.6 & 68.4 & 69.1 &  66.7 \\
    SQD-MapNet \cite{wang2024sqdmap} &  ECCV24 &  24 & - & - & - & - & 63.6 & 66.6 & 64.8 &  65.0 \\
     MapTRv2$^\dag$ \cite{liao2024maptrv2} &  IJCV24 &  24 & 34.2 & 39.3 & 36.2 & 36.6 & 58.1 & 60.8 & 62.3 &  60.4 \\
    MapQR \cite{liu2024mapqr} &  ECCV24 &  24 & 38.6 & 49.9 & 41.5 & 43.3 & 63.4 & 68.0 & 67.7 &  66.4 \\
    Mask2Map \cite{choi2024mask2map} &  ECCV24 &  24 & - & - & - & - & 70.6 & 71.3 & 72.9 &  71.6 \\
   \rowcolor{gray!20}MapTRv2+Ours  &  - &  24 & 34.7 & 43.2 & 39.1 & 39.0(+2.4) & 58.5 & 64.7 & 65.1 &  62.8(+2.4) \\
    \rowcolor{gray!20}MapQR+Ours &  - &  24 & 40.3 & 52.4 & 45.3 & 46.0(+2.6) & 65.2 & 70.8 & 70.3 &  68.8(+2.4) \\
    
    \rowcolor{gray!20}Mask2Map+Ours &  - &  24 & 55.6 & 62.2 & 59.4 & 51.2 & 70.7 & 72.3 & 74.9 &  72.6(+1.0) \\
    \hline
    VectorMapNet \cite{liu2023vectormapnet} &  ICML23&  110&  18.2 &  27.2 &  18.4 &  21.3 & 36.1 & 47.3 & 39.3 & 40.9 \\ 
    MapTR$^\ddag$ \cite{liaomaptr} &  ICLR23&  110& 31.4 & 40.5 & 35.5 & 35.8 &  
    55.8 & 60.9	& 61.1 & 59.3 \\
    BeMapNet \cite{qiao2023bemap}&  CVPR23&  110& 44.5 & 52.7 & 44.2 & 47.1 & 62.6&  66.7&  65.1&  64.8\\ 
    MapTRv2 \cite{liao2024maptrv2} &  IJCV24 &  110 & - & - & - & - & 68.1 &  68.3 &  69.7 &  68.7 \\ 
    MGMap \cite{liu2024mgmap} &  CVPR24 &  110 & 54.5 & 42.1 & 37.4 & 44.7 & 64.4 &  67.6 &  67.7 &  66.5 \\ 
    MapTRv2+SGQ \cite{liu2024mapqr} &  ECCV24 &  110 & 44.0 & 52.9 & 45.3 & 47.4 & 67.7 &  69.8 &  70.4 &  69.3 \\ 
    MapTRv2$^\dag$ \cite{liao2024maptrv2} &  IJCV24 &  110 & 43.4 & 48.2 & 42.9 & 44.9 & 67.5 & 68.3 & 69.1 & 68.3 \\
    \rowcolor{gray!20}MapTRv2+HLS (Ours) &  - &  110 & 44.8 & 50.0 & 46.3 & 47.0(+2.1) & 68.2 & 69.7 & 70.5 & 69.4(+1.1) \\
    \rowcolor{gray!20}MapTRv2+Ours &  - &  110 & 44.0 & 51.6 & 46.6 & 47.4(+2.5) & 68.6 & 71.3 & 71.2 & 70.4(+2.1) \\
    \Xhline{1.5pt}
    \end{tabular}
    }
    \caption{\textbf{Comparison to the state-of-the-art on nuScenes \textit{val} set with ResNet-50.} The $\dag$ is the reproduced result from public code. Our reproduced models trained on 4 NVIDIA GeForce RTX 3090 GPUs with batch size 16 and $3\times10^{-4}$ with cosine decay, which is half the learning rate at total batch size 32. So some reproduced results are low than the results with batch size 32 from paper. The numbers in brackets are corresponding mAP gains compared with reproduced baseline models under the same settings. The $\ddag$ means the results from released model. ``-" means that the corresponding results are not available. The metric of easy setting and hard setting are follow \cite{ding2023pivotnet}.
    }
    \label{tab:sota_nus}
\end{table*}

\subsection{Task Modulated Deformable Attention}

Sharing the same instance features may cause inherent conflicts across tasks, leading to sub-optimal task output results. For example, features in salient regions within instances may have rich semantic information for classification tasks, while features at instance boundaries may be more conducive to localization tasks. 
The shared paradigm in current decoder designs prevents decoders from obtaining more discriminative task-specific features to achieve further performance gains.

To deal with the above potential issue, we propose the Task Modulated Deformable Attention to obtain discriminative task-specific features. Specifically, we first double the channels of the initialized query $\mathbf{Q}$, with half of the channels responsible for classification and the other half for localization. Then, as with the conventional decoder, the queries utilize the self-attention layer for interactions between different queries. Next, we split the queries obtained from the self-attention layer to generate task-aware queries $\mathbf{Q}_{\mathrm{cls}}$ and $\mathbf{Q}_{\mathrm{loc}}$ as follows:
\begin{equation}
  \mathbf{Q}_{\mathrm{cls}}, \mathbf{Q}_{\mathrm{loc}}=\mathrm{Split}(\mathrm{SA}(\mathbf{Q})),
\end{equation}
where $\mathbf{Q}\in \mathbb{R}^{2C}$, $\mathbf{Q}_{\mathrm{cls}}\in \mathbb{R}^{C}$,$\mathbf{Q}_{\mathrm{loc}}\in \mathbb{R}^{C}$. $\mathrm{Split}(\cdot)$ represents the split operation for the channels. 

After obtaining the task-aware queries, we feed them to the cross-attention layer to obtain the task-specific features. It is worth noticing that we do not simply use two cross-attention layers (Deformable Attention module \cite{zhudeformable}) to process task-aware queries separately, but utilize our proposed Task Decoupled Deformable Attention module to process them. The standard deformable attention module can be simply formulated as:
\begin{align}
&\mathbf{A}=W_{a}{\mathbf{Q}}, \mathrm{\Delta} r=W_{p}{\mathbf{Q}}, \mathbf{V}=\mathrm{Samp}(W_v\mathbf{F}, r+\mathrm{\Delta} r),  \notag\\
&\mathrm{DeformableAttention}(\mathbf{Q},\mathbf{V}) = \mathrm{Softmax}(\mathbf{A})\mathbf{V},
\end{align}
where $\mathrm{\Delta} r$ are the offsets of reference points, $r$ are the reference points, $W_{a}$, $W_{p}$ and $W_{v}$ represent the linear projection. $\mathbf{A}$ are the attention weights of the sample features, Samp is a function that extracts the features corresponding to the location $(r+\mathrm{\Delta} r)$ by bilinear interpolation.

Our proposed Task Decoupled Deformable Attention module utilizes task-aware query to learn task-specific attention weights, while offset learning remains shared. The reason for this design is that we argue that 1) it is difficult to simultaneously learn task-specific weights and offsets using the task-aware query due to the large number of variables to be optimized, and 2) offset learning itself is more difficult to optimize because its optimization objective is unbounded. Our proposed Task Decoupled Deformable Attention module can be simply formulated as:
\begin{align}
&\mathbf{A_{\mathrm{cls}}}=W_{a}{\mathbf{Q}_{\mathrm{cls}}}, \mathbf{A_{\mathrm{loc}}}={W_{a}^{'}}{\mathbf{Q}_{\mathrm{loc}}} \notag\\
&\mathrm{\Delta} r={W_{p}}\mathrm{Cat}(\mathbf{Q}_{\mathrm{cls}}, \mathbf{Q}_{\mathrm{loc}}), \mathbf{V}=\mathrm{Samp}(W_v\mathbf{F}, r+\mathrm{\Delta} r),  \notag\\
&\hat{\mathbf{Q}}_{\mathrm{cls}},\hat{\mathbf{Q}}_{\mathrm{loc}} = \mathrm{Softmax}(\mathbf{A}_{\mathrm{cls}})\mathbf{V}, \mathrm{Softmax}(\mathbf{A}_{\mathrm{loc}})\mathbf{V},
\end{align}
where $\mathrm{Cat}$ represents the concatenation operation.

Next, we feed the task-specific features extracted by the cross-attention layer into different FFN networks, respectively. Finally, task-specific features are fed into the two task heads to output the predictions of the map element.

\section{Experiments}
\label{sec:experiments}

\subsection{Datasets}
\textbf{NuScenes.} NuScenes \cite{caesar2020nuscenes} dataset includes 1000 driving scenes recorded in Boston and Singapore. Each scene lasts about 20 seconds and is sampled at 2Hz to get 40 key-frame samples. For each sample, there are 6 RGB images from surrounding cameras that provide the full 360-degree field of view. Following previous methods, 700/150/150 split scenes are utilized for training/validation/testing, respectively. We report the performance of three categories, including the road boundary, the lane divider, and the pedestrian crossing, on the validation set. The ground truth in the NuScenes is annotated as the 2D vectorized map elements.

\noindent\textbf{Argoverse2.} Argoverse2 \cite{wilson2argoverse} dataset includes 1000 driving scenes recorded in six cities. Each scene lasts about 15 seconds and is sampled at 10Hz to get 150 key-frame samples. For each sample, there are 7 RGB images from surrounding cameras. Following previous methods, 700/150 split scenes with 110K/24K frames are used for training/validation.
We report the performance of the same categories as NuScenes on the validation set. The ground truth in Argoverse2 is annotated as the 3D vectorized map elements.

\begin{table*}[ht]
\setlength{\tabcolsep}{3.66pt}
\setlength{\abovecaptionskip}{0.1cm}
\setlength{\belowcaptionskip}{0.cm}
    \centering
    \begin{tabular}{c| c| c|c c c |c|c c c |c} 
    \Xhline{1.5pt}
    \multirow{2}{*}{Methods} &
    \multirow{2}{*}{Backbone} &
    \multirow{2}{*}{Epoch} &
    $\text{AP}_{ped.}$ & $\text{AP}_{div.}$ & $\text{AP}_{bou.}$ & mAP &
    $\text{AP}_{ped.}$ & $\text{AP}_{div.}$ & $\text{AP}_{bou.}$ & mAP \\
    \cline{4-11}
    & & & \multicolumn{4}{c|}{ $hard:\{0.2, 0.5, 1.0\}m$} & \multicolumn{4}{c}{ $easy:\{0.5, 1.0, 1.5\}m$} \\
    \hline
    MapQR \cite{liu2024mapqr} & SwinB  &  24&  41.4 &  52.3 &  43.2  &  45.6 & 67.5 & 72.1 & 70.5 & 70.1\\ 
    MapQR$^\dag$ \cite{liu2024mapqr} &  SwinB &  24 & 41.6 & 53.2 & 46.3 & 47.0 & 66.9 & 71.9 & 71.0 &  70.0 \\
    \rowcolor{gray!20}MapQR+Ours  & SwinB  &  24 & 43.2 & 54.6 & 47.9 & 48.6(+1.6) & 68.7 & 73.9 & 73.2 &  71.9(+1.9) \\
    \Xhline{1.5pt}
    \end{tabular}
    \caption{\textbf{Comparison to the state-of-the-art on nuScenes \textit{val} set with SwinB.} 
    The $\dag$ is the reproduced result from public code. 
    }
    \label{tab:sota_nus_swin}
\end{table*}

\begin{table*}[ht]
\setlength{\tabcolsep}{3.66pt}
\setlength{\abovecaptionskip}{0.1cm}
\setlength{\belowcaptionskip}{0.cm}
    \centering
    \begin{tabular}{c| c| c|c c c |c|c c c |c} 
    \Xhline{1.5pt}
    \multirow{2}{*}{Methods} &
    \multirow{2}{*}{Dim} &
    \multirow{2}{*}{Epoch} &
    $\text{AP}_{ped.}$ & $\text{AP}_{div.}$ & $\text{AP}_{bou.}$ & mAP &
    $\text{AP}_{ped.}$ & $\text{AP}_{div.}$ & $\text{AP}_{bou.}$ & mAP \\
    \cline{4-11}
    & & & \multicolumn{4}{c|}{ $hard:\{0.2, 0.5, 1.0\}m$} & \multicolumn{4}{c}{ $easy:\{0.5, 1.0, 1.5\}m$} \\
    \hline
    MapTRv2\cite{liao2024maptrv2} &   \multirow{9}{*}{3} & 6&  30.5 &  48.1 &  36.7  &  38.4 & 60.7 & 68.9 & 64.5 & 64.7\\ 
    MapTRv2$^\dag$ \cite{liao2024maptrv2} &   &  6 & 30.1 & 47.9 & 36.4 & 38.1 & 58.4 & 69.0 & 63.4 &  63.6 \\
    \rowcolor{gray!20}MapTRv2+Ours  &   &  6 & 32.5 & 49.7 & 40.5 & 40.9(+2.8) & 61.0 & 70.3 & 67.2 &  66.2(+2.6) \\
    MapQR \cite{liu2024mapqr} &   &  6 & 30.6 & 49.4 & 38.9 & 39.6 & 60.1 & 71.2 & 66.2 &  65.9 \\
    MapQR$^\dag$ \cite{liu2024mapqr} &   &  6 & 33.9 & 50.1 & 39.5 & 41.1 & 61.3 & 68.9 & 66.1 &  65.4 \\
   \rowcolor{gray!20} MapQR+Ours &   &  6 & 35.3 & 52.4 & 43.3 & 43.6(+2.5) & 62.9 & 70.9 & 68.4 &  67.4(+2.0) \\ \cline{1-1} \cline{3-11}
    MapTRv2$^\dag$ \cite{liao2024maptrv2} &   &  24 & 36.3 & 52.3 & 41.8 & 43.5 & 65.1 & 72.2 & 68.1 & 68.5 \\
    \rowcolor{gray!20}MapTRv2+Ours &   &  24 & 36.0 & 54.2 & 45.0 & 45.0(+1.5) & 64.7& 73.2 & 70.2 & 69.4(+0.9) \\
    MapQR$^\dag$ \cite{liu2024mapqr} &   &  24 & 36.7 & 54.5 & 44.0 & 45.1 & 63.3 & 71.0 & 69.3 &  67.9 \\
    \rowcolor{gray!20}MapQR+Ours &   &  24 & 37.6 & 55.6 & 46.2 & 46.5(+1.4) & 64.4 & 72.1 & 69.8 &  68.8(+0.9) \\
    \Xhline{1.5pt}
    \end{tabular}
    \caption{\textbf{Comparison to the state-of-the-art on Argoverse2 \textit{val} set with ResNet-50.} 
    The $\dag$ is the reproduced result from public code. The reproduced setting is same as the nuScenes. The numbers in brackets are corresponding mAP gains. 
    }
    \label{tab:sota_argo}
\end{table*}
\subsection{Metric} Following previous methods, the mean Average Precision (mAP) metric based on Chamfer Distance is chosen for evaluation. A prediction that satisfies a specific chamfer distance threshold is defined as a True-Positive (TP). Two different threshold sets (hard and easy) are adopted for a more comprehensive evaluation. The hard and easy settings correspond to \{0.2, 0.5, 1.0\}$m$ and \{0.5, 1.0, 1.5\}$m$, respectively. The prediction map range is $[-15.0m, 15.0m]$ for the $X$-axis and $[-30.0m, 30.0m]$ for the $Y$-axis.
\subsection{Implementation Details} We adopt the ResNet50 as the backbone to conduct experiments unless otherwise specified. The total batch size is 16 and the learning rate is $3\times10^{-4}$. All models are trained on 4 NVIDIA GeForce RTX 3090 GPUs unless otherwise specified. The training schedule is 24 epochs for NuScenes and 6 epochs for Argoverse2 unless otherwise specified. Other settings are the same as the baseline MapTRv2 \cite{liao2024maptrv2}.

\subsection{Main Results}
\noindent\textbf{Results on nuScenes.} For nuScenes \cite{caesar2020nuscenes}, different methods output 2D vectorized HD map elements for evaluation. 
Table \ref{tab:sota_nus} shows the comparison of the experimental results against the SOTA methods. Combining our components with MapTRv2, our proposed DAMap achieves 62.8 and 39.0 mAP, which are 2.4 and 2.4 mAP higher than our reimplemented MapTRv2 baseline with the 24 epochs setting. Since recent work MapQR \cite{liu2024mapqr} and Mask2Map \cite{choi2024mask2map} have shown good performance, we also chose them as the baselines to validate the generality of our method. Combining our components with MapQR, our proposed DAMap achieves 68.8 and 46.0 mAP, which are 2.4 and 2.6 mAP higher than our reimplemented MapQR baseline with the 24 epochs. Combining our components with Mask2Map, our proposed DAMap achieves 72.6 mAP, which is 1.0 mAP higher than Mask2Map baseline with the 24 epochs.

To evaluate the effectiveness of our method under a longer training schedule, we perform experiments under the 110 epochs training schedule. Our method achieves 70.4 and 47.4 mAP, which are 2.1 and 2.5 mAP higher than the baseline. Besides, we conduct the experiment of introducing only our HLS with DAFL into the baseline MapTRv2 \cite{liao2024maptrv2}. It is worth noting that our HLS with DAFL as the loss function does not introduce parameters and computational effort in the inference stage. Our proposed HLS with DAFL achieves 69.4 and 47.0 mAP, which are 1.1 and 2.1 mAP higher than the baseline MapTRv2. 

In addition, our DAMap can consistently obtain performance gains even with a stronger Swin-Transformer \cite{liu2021swin} backbone. As shown in Table \ref{tab:sota_nus_swin}, when using SwinB as the backbone, our DAMap achieves 71.9 and 48.6 mAP, which are 1.9 and 1.6 mAP higher than the baseline.
These results validate the effectiveness and generalizability of our method on different metrics, baselines, schedules, and backbones.

\noindent\textbf{Results on Argoverse2.} For Argoverse2 \cite{wilson2argoverse}, different methods output 3D vectorized HD map elements for evaluation. We also evaluate our proposed DAMap on the Argoverse2 val dataset under different metrics (hard and easy), baselines (MapTRv2 \cite{liao2024maptrv2} and MapQR \cite{liu2024mapqr}), and schedules (6 and 24 epochs). Table \ref{tab:sota_argo} shows the comparison of the experimental results. Our proposed DAMap consistently obtains performance improvement under all settings. Specifically, under the 6 epochs schedule, our method outperforms MapTRv2 by 2.6 and 2.8 mAP and MapQR by 2.0 and 2.5 mAP, respectively. Under the 24 epochs schedule, our method outperforms MapTRv2 by 0.9 and 1.5 mAP and MapQR by 0.9 and 1.4 mAP, respectively. These results validate the effectiveness and generalizability of our method on the different datasets.

\begin{table}[t]
\setlength{\tabcolsep}{1.5pt}
\setlength{\abovecaptionskip}{0.1cm}
\setlength{\belowcaptionskip}{0.cm}
    \centering
    \begin{tabular}{c| c c c|c} 
    \Xhline{1.5pt}
   Methods   &$\text{AP}_{ped.}$ & $\text{AP}_{div.}$ & $\text{AP}_{bou.}$  & mAP \\ \hline
MapTRv2$\dag$\cite{liao2024maptrv2}        & 12.1  & 26.3  & 42.2   & 26.9  \\ 
 \rowcolor{gray!20}MapTRv2+Ours & 12.3 & 27.6 & 45.8 &  28.6(+1.7) \\ 
 StreamMapNet\cite{yuan2024streammapnet} & 32.2  & 29.3  & 40.8   & 34.1 \\
 \rowcolor{gray!20}StreamMapNet+Ours & 35.1  & 27.8  & 43.8   & 35.6(+1.5) \\
\Xhline{1.5pt}
    \end{tabular}
    \caption{\textbf{Comparison to the baseline methods on new nuScenes split \textit{val} set under the ResNet50 backbone.} 
    The $\dag$ is the reproduced result from public code. 
    }
    \label{tab:sota_nus_newsplit}
\end{table}
\noindent\textbf{Results on New nuScenes Split.} To ensure a fair comparison, we also perform experiments on the new nuScenes split proposed by StreamMapNet \cite{yuan2024streammapnet}. As shown in Table \ref{tab:sota_nus_newsplit}, our proposed DAMap demonstrates superior performance over baseline methods across all categories in the new nuScenes split. The results show the effectiveness of our method on different dataset splits.

\begin{table}[t]
\setlength{\tabcolsep}{1.8pt}
\setlength{\abovecaptionskip}{0.1cm}
\setlength{\belowcaptionskip}{0.cm}
    \centering
    \scalebox{0.95}{
    \begin{tabular}{c|c| c c c c|c} 
    \Xhline{1.5pt}
    Methods &
    Data. &
    $\text{AP}_{ped.}$ & $\text{AP}_{div.}$ & $\text{AP}_{bou.}$ & $\text{AP}_{cen.}$ & mAP \\
    \hline
    MapTRv2\cite{liao2024maptrv2}   & nuS&   50.1 & 53.9 & 58.8 & 53.1 & 54.0\\ 
    MapTRv2$^\dag$\cite{liao2024maptrv2}   & nuS& 47.7 & 52.8 & 59.2 & 53.2&  53.2 \\
    \rowcolor{gray!20}MapTRv2+Ours   & nuS& 51.0 & 60.6 & 64.5 & 57.2 &  58.3 \\ \hline
    MapTRv2$^\dag$\cite{liao2024maptrv2}   & Arg2& 51.6 & 62.1 & 61.3 & 58.9&  58.5 \\ 
    \rowcolor{gray!20}MapTRv2+Ours   & Arg2& 53.5 & 68.6 & 64.7 & 63.3 &  62.5 \\
    \Xhline{1.5pt}
    \end{tabular}}
    \caption{\textbf{Comparison to the  the baseline methods on nuScenes and Argoverse2 \textit{val} set with Centerline category under the ResNet50.} 
    The $\dag$ is the reproduced result from public code. 
    }
    \label{tab:sota_nus_center}
\end{table}
\noindent\textbf{Results with Centerline Category.} To investigate the effectiveness for more categories, we conduct experiment of our method with the Centerline category. As shown in Table \ref{tab:sota_nus_center}, our DAMap can obtain performance gains even with the Centerline category. Specifically, our DAMap achieves 58.3 mAP and 62.5 mAP in nuScenes and Argoverse2, respectively, which is 5.1 mAP and 4.0 mAP higher than the reproduced baseline.
These results further validate the effectiveness and generalizability of our method.

\begin{table}[t]
\setlength{\abovecaptionskip}{0.1cm}
\setlength{\belowcaptionskip}{0.cm}
\centering
\scalebox{0.9}{
\begin{tabular}{ccc|cccc}
\Xhline{1.5pt}
DAFL & HLS & TMDA & $\text{AP}_{ped.}$ & $\text{AP}_{div.}$ & $\text{AP}_{bou.}$ & mAP  \\ \hline
  -   &   -  &   -  & 58.1  & 60.8  & 62.3  & 60.4 \\ \hline
\checkmark    &     &     & 58.6  & 61.1  & 63.1  & 60.9 \\
\checkmark    & \checkmark   &     & 58.5  & 62.9  & 63.4  & 61.6 \\
     &     & \checkmark   & 58.5  & 63.1  & 63.2  & 61.6 \\
\checkmark    & \checkmark   & \checkmark  & 58.5  & 64.7  & 65.1  & 62.8 \\ 
\Xhline{1.5pt}
\end{tabular}}
\caption{\textbf{Ablations on each component of our DAMap.} "DAFL", "HLS" and "TMDA" represent Distance-aware Focal Loss, Hybrid Loss Scheme and Task Modulated Deformable Attention, respectively.}
\vspace{-1.2em}
\label{tab:components}
\end{table}
\subsection{Ablation Study}
\textbf{Ablation on Each Component.} To analyze the effect of each component in our DAMap, we gradually apply Distance-aware Focal Loss (DAFL), Hybrid Loss Scheme (HLS), and Task Modulated Deformable Attention (TMDA) to the baseline. As shown in Table \ref{tab:components}, the DAFL produces 0.5 mAP gains compared to the baseline. The result shows that introducing the localization quality into the Focal Loss is beneficial to the model. The HLS with DAFL improves the performance from 60.4 mAP to 61.6 mAP. This result suggests that our HLS can take greater advantage of the DAFL. The TMDA brings 1.2 mAP gains compared to the baseline. When all components are introduced into the baseline, it achieves 62.8 mAP with 2.4 mAP gains. Hyperparameter experiments on these components and runtime analysis can be found in the Supplementary Material.
The results show that these components are complementary and address different issues in the HD map task.

\noindent\textbf{Ablation on TMDA Design.} To validate the effectiveness of our TMDA design, we conduct experiments on other designs with the same complexity. As shown in Table \ref{tab:tmda}, 
other designs can obtain a slight gain. For example, the design (Setting 1) of task-specific sampling offsets and attention weights generated by the task-aware query can bring 0.3 mAP gains. This design is equivalent to having a separate cross-attention layer for each of the classification and localization tasks. 
Our design (Setting 4) of task-agnostic sampling offsets generated by task-agnostic query and the task-specific attention weights generated by task-aware query can achieve 1.2 mAP gains. The results show that our method outperforms the other designs. In addition, we also show the attention weights visualization of TMDA in the Supplementary Material.

\begin{table}[t]
\setlength{\abovecaptionskip}{0.1cm}
\setlength{\belowcaptionskip}{0.cm}
\centering
\begin{tabular}{c|cccc|c}
\Xhline{1.5pt}
Setting   &$\text{AP}_{ped.}$ & $\text{AP}_{div.}$ & $\text{AP}_{bou.}$  & mAP &Para.\\ \hline
-         & 58.1  & 60.8  & 62.3   & 60.4 &40M \\ \hline
 1 & 57.5  & 61.8  & 62.9   & 60.7 &52M\\
 2 & 57.4  & 63.0  & 62.3   & 60.9 &52M\\
 3 & 57.0  & 62.6  & 61.7   & 60.4 &52M\\
4 (Ours)      & 58.5  & 63.1  & 63.2    & 61.6 &52M\\ 
\Xhline{1.5pt}
\end{tabular}
\caption{\textbf{Ablations on our TMDA with other designs.} 1: task-specific sampling offsets and attention weights are generated by task-aware query. 2: task-specific sampling offsets and attention weights are learned by task-agnostic query (concatenated task-aware query). 3: task-specific sampling offsets and task-agnostic attention weights are learned by task-agnostic query.}
\label{tab:tmda}
\end{table}

\begin{figure}[t]
\setlength{\abovecaptionskip}{0.1cm}
\setlength{\belowcaptionskip}{-0.5cm}
 \centering
\includegraphics[width=0.98\linewidth]{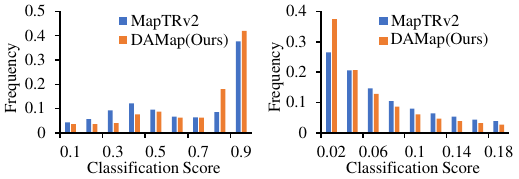}
 \caption{Left: Frequency of the classification scores on the True Positive with or without our method. Right: Frequency of the classification scores on the False Positive with or without our method.}
 \label{fig:cls_statistic}
\end{figure}

\noindent\textbf{Quantitative Analysis of Classification Score.} To further illustrate that our method can produce high quality predictions, we count the classification scores of the predictions. As shown in Figure \ref{fig:cls_statistic}, 
our method improves the percentage of predictions with high classification score for True Positive (Left) and the percentage of predictions with low classification score for False Positive (Right). 
The results show that our method can encourage positive samples and suppress negative samples. This phenomenon is further evidence that our method can produce high quality predictions. Quantitative analysis of localization quality can be found in the Supplementary Material.

\section{Conclusion}
In this paper, we analyze two inherent defects in current methods that limit the high quality predictions. To address these problems, we propose a novel map construction method named DAMap, which consists of three components: Distance-aware Focal Loss (DAFL), Hybrid Loss Scheme (HLS) and Task Modulated Deformable Attention (TMDA). The DAFL is introduced to assign appropriate labels for one-to-many matching samples. The TMDA is proposed to obtain discriminative task-specific features. Furthermore, the HLS is proposed to better utilize the advantages of the DAFL. Extensive experiments conducted on the nuScenes and Argoverse2 benchmarks under different metrics, baselines, splits, backbones, and schedules validate the effectiveness of our method.
\section*{Acknowledgements}
This work is supported by National Natural Science Foundation of China (NSFC) under Grant 62088102. 
We also appreciate Yufeng Hu for his suggestions. We thank all the anonymous reviewers and Area Chairs for their constructive and helpful comments, which have significantly improved the quality of the paper.

{
    \small
    \bibliographystyle{ieeenat_fullname}
    \bibliography{main}
}
\clearpage
\setcounter{page}{1}
\maketitlesupplementary
\appendix
In the supplementary material, we first conduct more ablation studies. Then, we report the runtime analysis of our DAMap. Finally, we present the visual analysis of map element predictions on the nuScenes val set.

\section{More Ablation Study}
\begin{table}[ht]
\setlength{\abovecaptionskip}{0.1cm}
\setlength{\belowcaptionskip}{0.cm}
\centering
\begin{tabular}{c|ccccc}
\Xhline{1.5pt}
Hybrid Number & $\text{AP}_{ped.}$ & $\text{AP}_{div.}$ & $\text{AP}_{bou.}$  & mAP \\ \hline
-              & 58.1  & 60.8  & 62.3   & 60.4 \\ \hline
0             & 58.6  & 61.1  & 63.1    & 60.9 \\
1             & 58.5  & 62.9  & 63.4   & 61.6 \\
2             & 58.7  & 60.8  & 63.0   & 60.8 \\
3             & 56.8  & 62.1  & 62.6   & 60.5 \\ 
\Xhline{1.5pt}
\end{tabular}
\caption{Ablations on the hybrid number in Hybrid Loss Scheme.}
\label{tab:hls}
\end{table}
\noindent\textbf{Ablation on Hybrid Number in HLS.} We analyze the effect of the hybrid number in HLS. The results are shown in Table \ref{tab:hls}. When using the Focal Loss \cite{lin2017focal} in the full decoder layer, the baseline performance is 60.4 mAP. When using our proposed DAFL in the full decoder layer, the performance of the model is 60.9 mAP. When Focal Loss is introduced as the hybrid loss in the first decoder layer, the model achieves 61.6 mAP. The result shows that using Focal Loss to obtain higher quality predictions than utilizing DAFL has better results. When we continue to introduce the Focal Loss into more decoder layers, the performance improvement of the model decreases. Thus, we adopt one decoder layer with Focal Loss in all our experiments.

\begin{table}[ht]
\setlength{\abovecaptionskip}{0.1cm}
\setlength{\belowcaptionskip}{0.cm}
\centering
\begin{tabular}{c|ccc|c}
\Xhline{1.5pt}
$\lambda$   &$\text{AP}_{ped.}$ & $\text{AP}_{div.}$ & $\text{AP}_{bou.}$  & mAP \\ \hline
 1.0 & 58.5 & 64.7 & 65.1 &  62.8 \\ 
 0.8 & 59.6  & 64.4  & 64.5   & 62.9 \\
 1.2 & 58.3  & 64.4  & 64.7   & 62.5 \\
\Xhline{1.5pt}
\end{tabular}
\caption{Ablation on the hyper-parameter $\lambda$ in the DAFL.}
\label{tab:hyper}
\end{table}
\noindent\textbf{Ablation on Hyper-parameter $\lambda$.}
The $\lambda$ is a hyper-parameter to adjust the distribution of loss to quality. To investigate the effect of the hyper-parameter $\lambda$ in the DAFL, we conduct experiment of our method with different $\lambda$ values. As shown in Table \ref{tab:hyper}, setting different $\lambda$ values has a small effect (0.3 margin) on the performance of the model. We set $\lambda$ to 1.0 in all MapTR experiments.

\begin{figure}[t]
\setlength{\abovecaptionskip}{0.1cm}
\setlength{\belowcaptionskip}{-0.2cm}
 \centering
\includegraphics[width=0.98\linewidth]{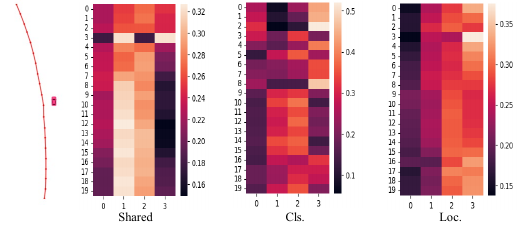}
 \caption{Example of learned attention weights for points in an instance when task-shared and task modulated.}
 \label{fig:att_vis}
\end{figure}

\noindent\textbf{Visualization of Attention Weights.} To further understand that classification and localization tasks often have different feature preferences, we show the visualization of attention weights in Figure \ref{fig:att_vis}. It can be seen that the weight distribution of both tasks is different. The weight distribution for each task is different from the weight distribution when tasks are shared. These results show that our Task Modulated Deformable Attention can obtain different weights for different tasks.
\begin{figure}[t]
\setlength{\abovecaptionskip}{0.1cm}
\setlength{\belowcaptionskip}{-0.2cm}
 \centering
\includegraphics[width=0.98\linewidth]{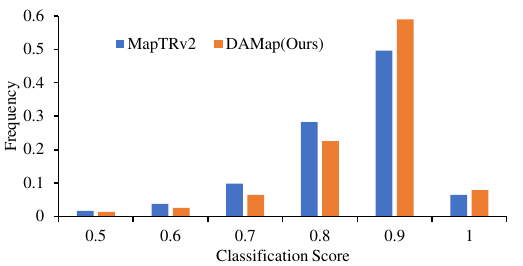}
 \caption{Frequency of the localization quality on the high classification score (0.9) with or without our method.}
 \label{fig:loc_statistic}
\end{figure}

\noindent\textbf{Quantitative Analysis of Localization Quality.} To further illustrate that our method can produce high quality predictions, we count the localization quality of the predictions under the 0.9 classification score. As shown in Figure \ref{fig:loc_statistic}, our method improves the percentage of predictions with high localization quality under high classification scores. 
The results show that our method can encourage high classification samples to obtain more accurate localization. This phenomenon is further evidence that our method can produce high quality predictions.

\begin{table}[ht]
\setlength{\abovecaptionskip}{0.1cm}
\setlength{\belowcaptionskip}{0.cm}
\centering
\begin{tabular}{c|c|c|c}
\Xhline{1.5pt}
Method & Para. & FPS  & mAP \\ \hline
MapTRv2\cite{liao2024maptrv2}              & 40M  & 14.1   & 68.7 \\ 
MapTRv2$^\dag$\cite{liao2024maptrv2}             & 40M  & 12.6   & 68.3 \\
MapTRv2+HLS(Ours)               & 40M  & 12.6   & 69.4 \\
DAMap(Ours)               & 52M  & 12.1   & 70.4 \\
\Xhline{1.5pt}
\end{tabular}
\caption{The analysis of Parameters and FPS of our DAMap. MapTRv2$^\dag$ is the inference speed in our environment.}
\label{tab:infer}
\end{table}
\section{Runtime Analysis}
We analyze the Parameters and FPS of our DAMap, as can be seen from Table \ref{tab:infer}. MapTRv2$^\dag$ is the inference speed in our environment, slightly lower than the MapTRv2 paper due to hardware differences (such as CPU). All FPS are the average inference speed by repeating the test 3 times on the NVIDIA RTX 3090Ti GPU. Our approach adds only a small amount of inference speed. Furthermore, our method does not increase the inference speed when equipped only with our HLS.

\section{Visualization of Map Element Predictions}
\begin{figure*}[ht]
\setlength{\abovecaptionskip}{0.1cm}
\setlength{\belowcaptionskip}{-0.2cm}
 \centering
\includegraphics[width=0.98\linewidth]{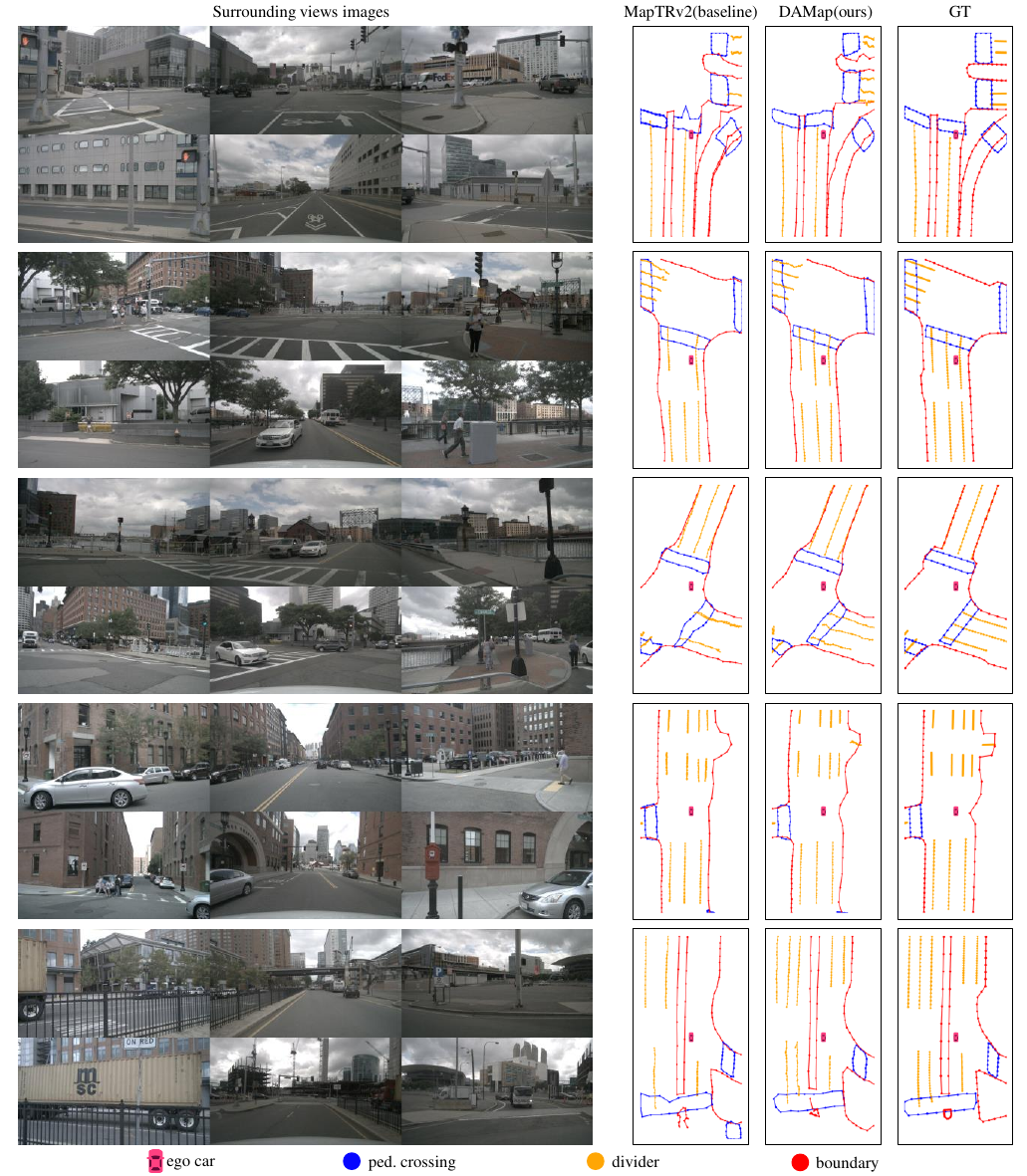}
 \caption{Comparison of map element predictions between our method and baseline with ResNet-50 and 24 epochs on the nuScenes val set. The score threshold is set to 0.4. }
 \label{fig:vis1_results}
\end{figure*}

\begin{figure*}[ht]
\setlength{\abovecaptionskip}{0.1cm}
\setlength{\belowcaptionskip}{-0.2cm}
 \centering
\includegraphics[width=0.98\linewidth]{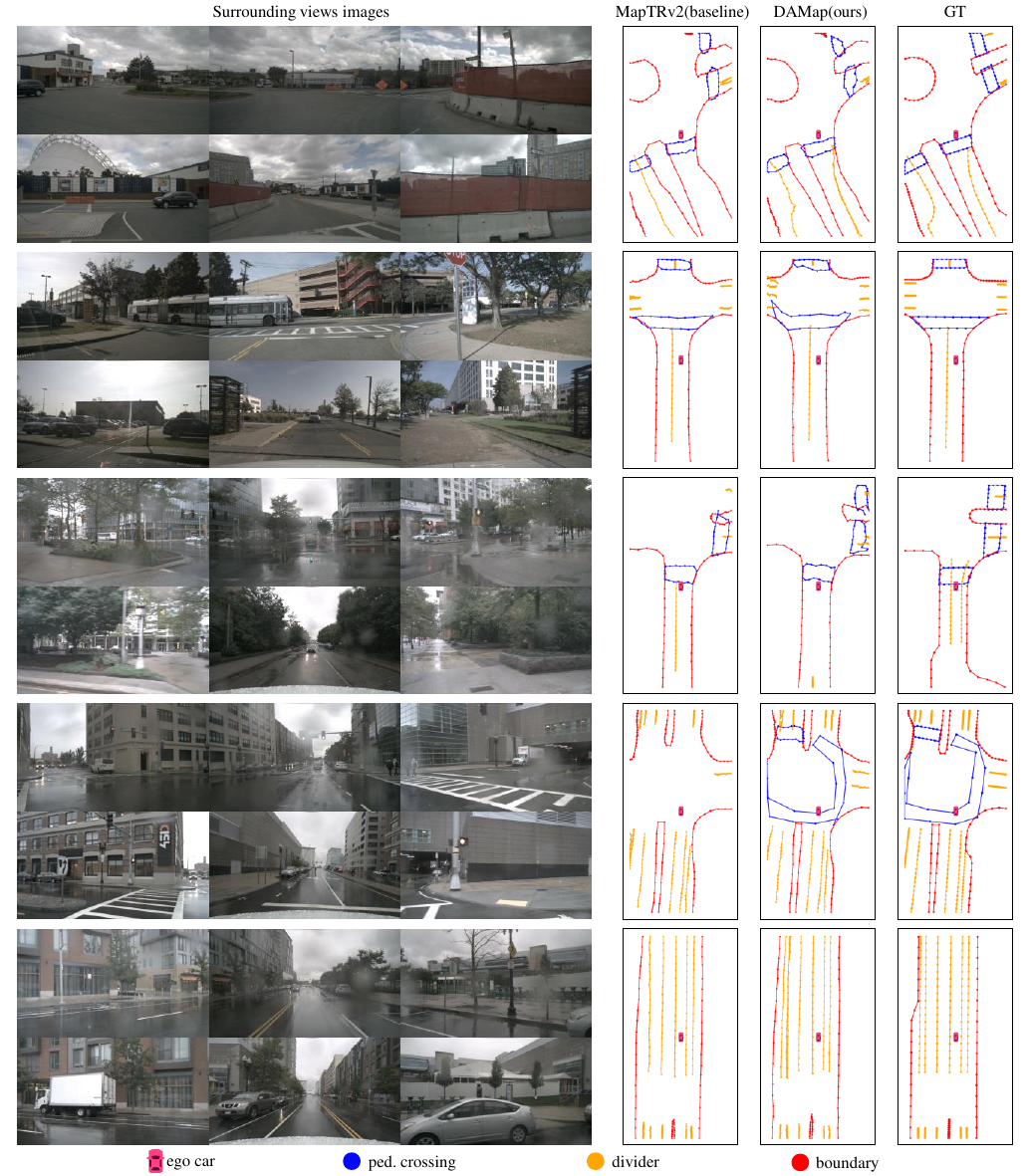}
 \caption{Comparison of map element predictions between our method and baseline with ResNet-50 and 24 epochs on the nuScenes val set. The score threshold is set to 0.4.}
 \label{fig:vis2_results}
\end{figure*}

As shown in Figure \ref{fig:vis1_results} and Figure \ref{fig:vis2_results}, we provide comparisons of map element predictions with the score threshold of 0.4 between the baseline and our method on the nuScenes val set. It can be seen that our method can output more map elements overlooked by the baseline under the same score threshold. This result shows that our method can achieve higher classification and higher localization results, further validating our motivation.


\end{document}